\documentclass[9pt]{article}
\usepackage{spconf,amsmath,graphicx}
\usepackage{hyperref} 
\usepackage{booktabs}
\usepackage{amssymb,xcolor,stackengine,graphicx}
\usepackage{multirow}

\title{Cascading and Direct Approaches to Unsupervised \\Constituency Parsing on Spoken Sentences}
\name{Yuan Tseng$^{1}$ \qquad Cheng-I Jeff Lai$^{2}$ \qquad Hung-yi Lee$^{1}$}
  
\address{$^{1}$ National Taiwan University \qquad $^{2}$ MIT CSAIL 
\\ \texttt{r11942082@ntu.edu.tw} }

\begin{document}
\ninept
\maketitle
%
\begin{abstract}
    Past work on unsupervised parsing is constrained to written form.
    In this paper, we present the first study on \emph{unsupervised spoken constituency parsing} given unlabeled spoken sentences and unpaired textual data. 
    The goal is to determine the spoken sentences' hierarchical syntactic structure in the form of constituency parse trees, such that each node is a span of audio that corresponds to a constituent.
    We compare two approaches: (1) cascading an unsupervised automatic speech recognition (ASR) model and an unsupervised parser to obtain parse trees on ASR transcripts, and
    (2) direct training an unsupervised parser on continuous word-level speech representations. 
    This is done by first splitting utterances into sequences of word-level segments, and aggregating self-supervised speech representations within segments to obtain segment embeddings.
    We find that separately training a parser on the unpaired text and directly applying it on ASR transcripts for inference produces better results for unsupervised parsing.
    Additionally, our results suggest that accurate segmentation alone may be sufficient to parse spoken sentences accurately. 
    Finally, we show the direct approach may learn head-directionality correctly for both head-initial and head-final languages without any explicit inductive bias. 
\end{abstract}
%
\section{Introduction}
\label{sec:intro}
Unsupervised constituency parsing is a long-standing research challenge in natural language processing \cite{10.5555/864689,klein-manning-2004-corpus,bod-2006-subtrees,shen2018neural} that aims to automatically determine the syntactic constituent structure of sentences without access to any training labels.
It sheds light on how children are able to learn high-level linguistic information, such as syntax and grammar, without expert supervision.
Additionally, constituency parse trees have also been shown to improve and provide greater interpretability to a variety of downstream tasks such as semantic role labeling \cite{strubell2018linguistically}, word representation learning \cite{kuncoro2020syntactic}, and speech synthesis \cite{Guo2019,tyagi20_interspeech,song2021}. 

\begin{figure*}[!htp]
  \centering
  \includegraphics[width=0.65\linewidth]{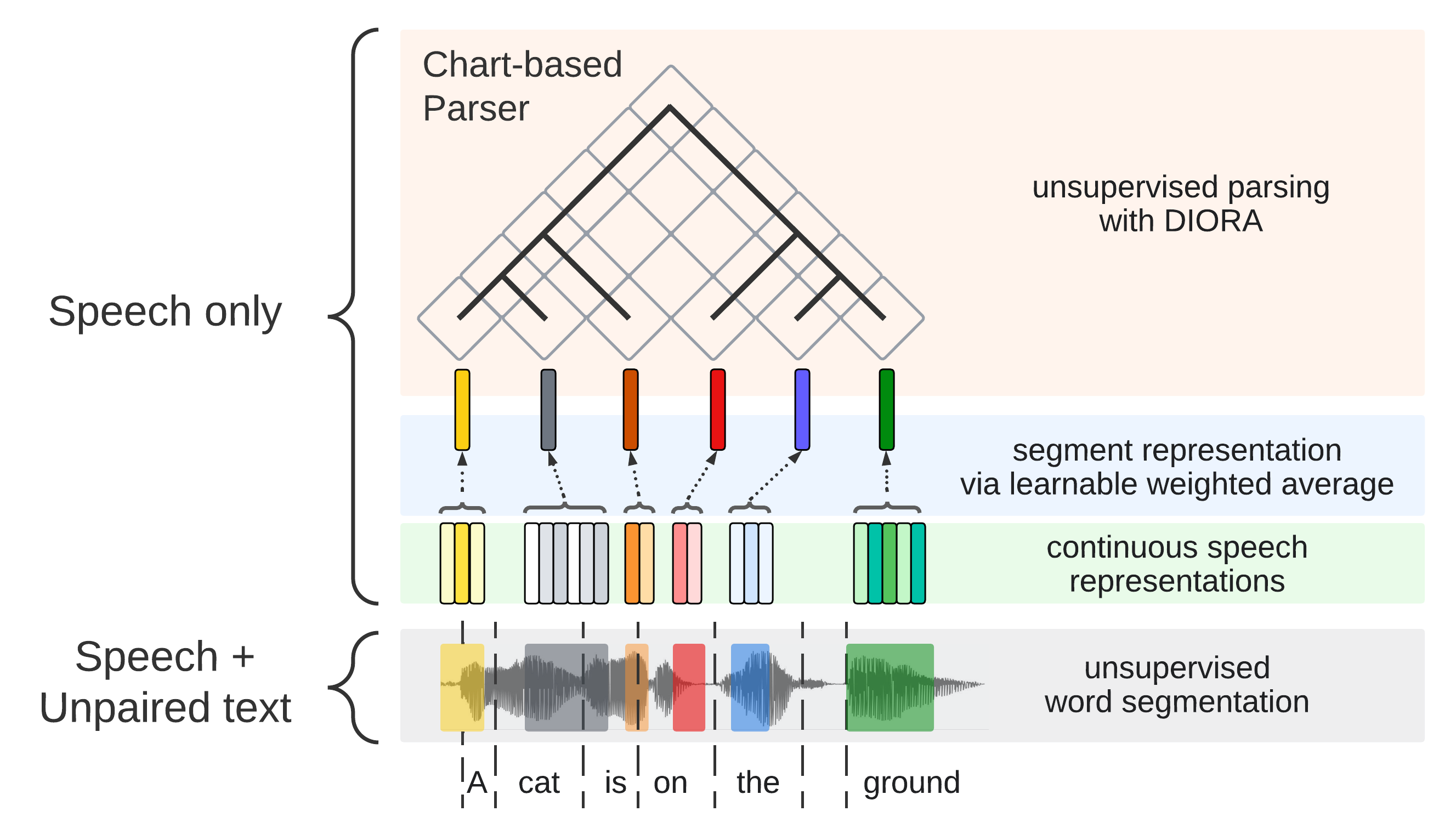}
  \caption{
  Diagram of our proposed direct approach to unsupervised spoken constituency parsing, using only raw speech and unpaired text.
  Textual transcripts of the input sentence are only shown for illustrative purposes.
  }
  \label{fig:weight_line}
\end{figure*}

To the best of our knowledge, syntactic parsing on speech was done exclusively in a supervised manner, using paired text transcripts and syntactic labels for training. 
However, these approaches cannot be applied to low-resource languages without paired data. 
This motivates us to explore a more realistic setting where only raw speech and limited unpaired textual data are available.

With no speech-text pairs or tree-text pairs, the first approach to unsupervised spoken constituency parsing is to cascade unsupervised ASR with an unsupervised parser. We compare training the parser on limited unpaired text and ASR transcripts and find that training on ASR transcripts does not help the model parse ASR transcripts of the same domain.

We also propose a framework to directly parse spoken input without any intermediate textual form.
First, we split an utterance into word-level segments, and transform each segment into a continuous embedding. Then we directly use this sequence of segment embeddings as input for our unsupervised parser.
We refer to this as the direct approach.

\vspace{1mm} \noindent \textbf{Contributions.}
(1) We perform the first investigation of unsupervised constituency parsing on spoken sentences using only raw speech and unpaired text. 
(2) We demonstrate that for parsing ASR transcripts, training on limited unpaired text is still better than training on ASR transcripts, and we quantify the effects of ASR errors on unsupervised parsing.
(3) We propose a framework to directly parse continuous speech without intermediate lexical unit discovery.
(4) We show that our direct approach may induce parse trees with the correct branching direction for different spoken languages.

\section{RELATED WORK}
\label{sec:format}
\subsection{Unsupervised Constituency Parsing}
\label{ssec:subhead}
Previous studies in unsupervised constituency parsing focus on obtaining constituency tree structures from large unannotated text corpora, usually by encouraging neural models to follow syntactic structure \cite{shen2018neural,shen2018ordered,kim2019unsupervised}, or parameterizing linguistic models with neural networks \cite{drozdov-etal-2019-unsupervised-latent,zhu-etal-2020-return,yang-etal-2021-neural}.

A recent line of work in visually-grounded grammar induction leverages paired images to improve unsupervised constituency parsing \cite{shi-etal-2019-visually,zhao2020visually,wan2021unsupervised}. We consider AV-NSL \cite{anonymous2023textless} in particular to be most relevant to our work, as they extend this approach to audio-visual learning and attempt to learn constituency parse trees from raw speech and image pairs. Unlike AV-NSL, our work does not rely on paired visual grounding data.

\subsection{Parsing Speech with Supervision}
\label{ssec:subhead}
Past works on syntactic parsing of speech address topics such as disfluency detection \cite{charniak2001edit,honnibal2014joint}, or incorporating prosodic cues \cite{kahn2005effective,dreyer07_interspeech}. However, most previous works require oracle transcripts, which is an unrealistic setting.
Yoshikawa et al. \cite{yoshikawa-etal-2016-joint} shows that it is possible to build a supervised dependency parser that jointly detect disfluencies and ASR errors, and Pupier et al. \cite{pupier22_interspeech} build an end-to-end supervised dependency parser for French that jointly predicts transcription and dependency tree from raw speech signals. Both works show an improvement over cascading baseline systems.

Additionally, prosodic features are shown to be closely related to syntax \cite{GROSJEAN197958,price1991use} and beneficial for both constituency parsing \cite{huang-harper-2010-appropriately,tran2018parsing} and dependency parsing \cite{ghaly2020using}, even under the presence of ASR errors \cite{tran21_interspeech}. These works motivate us to explore ways of using speech features to improve unsupervised constituency parsing of spoken data.

\subsection{Unsupervised Spoken Language Modeling}
\label{ssec:subhead}
Unsupervised spoken language modeling \cite{lakhotia-etal-2021-generative} aims to learn a spoken language model that simultaneously learns different levels of linguistic structure from raw speech signals with little or no textual data. 
The ZeroSpeech Challenge 2021 \cite{nguyen2020zero,dunbar2021zero} proposes to evaluate such a model at the acoustic, lexical, syntactic, and semantic levels. They find that while current spoken language models excel at the acoustic and lexical levels, higher levels of linguistic structure are much more difficult to model. They only require their models to be able to determine how grammatical a sentence is, while our work aims to solve the more challenging problem of producing the exact constituency structure of a sentence.

\vspace{-1mm}
\section{METHOD}
\label{sec:pagestyle}
\vspace{-1mm}
\subsection{Background}
\label{ssec:subhead}

Constituency parsing is usually formulated under the binary setting in order to reduce computation complexity. \cite{10.5555/1214993}
This setting entails that for a sentence with $n$ words $\{{x}_{1}, {x}_{2}, ...,  {x}_{n}\}$, each constituent spanning $x_{i:j}$ is composed of two constituents spanning $x_{i:k}$ and $x_{k+1:j}$ for some $k$ such that $i \leq k < j$. 
Our unsupervised parser follows the chart-based Deep Inside-Outside Recursive Autoencoder (DIORA) framework proposed by \cite{drozdov-etal-2019-unsupervised-latent} to produce binary parse trees without using any syntactically labeled data.  

\vspace{-2mm}
\subsubsection{Chart-based Constituency Parsing}
\label{sssec:chartbased}
Chart-based parsers find the optimal tree out of all valid binary parse trees by filling the upper-triangular portion of an $n \times n$ chart with a score $s_{i,j}$ for each cell. For $1 \leq i < j \leq n $, the score in cell $(i,j)$ represents how likely the span $x_{i:j}$ is a constituent. The CKY dynamic programming algorithm \cite{kasami1966efficient,younger1967recognition} is then used to determine the parse tree with the highest total score. 

\vspace{-2mm}
\subsubsection{Unsupervised Parser Architecture: DIORA}
\label{sssec:diora}
DIORA consists of an encoder and a decoder, and operates similarly to masked language models. 
The framework recursively encodes all but one of the words from the input sentence as a vector, and optimizes that vector to reconstruct the missing word. 
The core assumption is that the most efficient weights to produce such an encoding can be used as scores for chart-based constituency parsing.
DIORA initially represents the input sentence with pretrained ELMo character embeddings, but subsequent work \cite{wan2021unsupervised} shows that the framework can also be used with randomly initialized word embeddings.

\subsection{Cascading Parsing with Unsupervised ASR}
\label{ssec:subhead}
A straightforward approach to unsupervised spoken constituency parsing is to obtain word-level transcripts from unsupervised ASR, then represent each word with a randomly initialized vector. 
An unsupervised parser can then produce parse trees using these vector sequences as input. 

Our unsupervised ASR system adopts the wav2vec-U framework \cite{baevski2021unsupervised}.
wav2vec-U first phonemizes unpaired text data, then performs a series of preprocessing steps on unlabeled speech to produce higher-level features with length similar to phoneme sequences. 
They use adversarial training to train a model to predict phoneme sequences from speech features. 
A weighted finite-state transducer (WFST) trained on the unpaired text data is then used to decode the output into words. Further improvements are be obtained through Hidden Markov Model (HMM) self-training.
The phoneme output of the HMM achieves a lower phone error rate and can be decoded into more accurate word-level transcripts. 

\subsection{Direct Parsing on Speech Segments}
\label{ssec:subhead}

The direct parsing approach extends the DIORA framework by training on continuous word-level speech embeddings instead of ELMo embeddings. 
Using continuous segment representations allows our parser to benefit from continuous information in speech, in comparison to the discretization approach recently proposed in spoken language modeling \cite{lakhotia-etal-2021-generative}. 
This design choice is supported by AV-NSL \cite{anonymous2023textless}, which finds that continuous segment representations outperforms discrete representations for audio-visual parsing.

From each spoken utterance, we prepare (1) frame-level features, and (2) word-level segments.
Frame-level features can extracted from a pretrained self-supervised speech model such as XLSR-53 \cite{baevski2021unsupervised}, and word-level segments can be determined with unsupervised word segmentation models \cite{bhati21_interspeech,kamper2022word}.
Mirroring AV-NSL, we represent each segment with a continuous embedding parameterized by a simple weighted average of frame-level features. 
Weights are determined by a learnable two-layer MLP that is jointly optimized with the parser. 
This sequence of word-level segment embeddings is then directly used as input for our parser, then jointly optimized with the reconstruction loss proposed in DIORA \cite{drozdov-etal-2019-unsupervised-latent}.

\section{Experiments}
\label{sec:exps}
\subsection{Datasets, Preprocessing, and Hyperparameters}
\label{ssec:subhead}
Experiments are mainly conducted on the SpokenCOCO dataset \cite{hsu-etal-2021-text}, 
a 742h English read-speech dataset produced by 2.3k speakers reading the captions in MSCOCO \cite{lin2014microsoft}. 
Each image in MSCOCO corresponds to 5 captions on average. 
Following \cite{zhao2020visually}, we use the spoken captions of the 83k/5k/5k image split for training, validation, and testing respectively. The textual captions of the remaining 31k images are used as unpaired text data for unsupervised ASR. 
We focus on the more practical setting of unsupervised spoken constituency parsing using speech and unpaired text data only, hence we do not utilize the image data. 

Additional experiments in Korean are done on the Zeroth-Korean corpus\footnote{\href{https://github.com/goodatlas/zeroth}{https://github.com/goodatlas/zeroth}}, which contains 51.6hrs of audio spoken by 105 speakers for training, and 1.6hrs by 10 different speakers for testing.
We use the utterances of 10 speakers in the original training set for validation.

We note that due to a lack of labelled speech data, we are limited to experimenting on high-resource languages.
Following \cite{shi-etal-2019-visually, zhao2020visually}, ground-truth parse trees are obtained from the outputs of an off-the-shelf parser \cite{kitaev-klein-2018-constituency} on the normalized text captions. 
Punctuation is removed from the trees, and we run forced alignment using the Montreal Forced Aligner \cite{mcauliffe17_interspeech} to obtain oracle word boundaries.

For all experiments, we use the same hyperparameters as the randomly initialized DIORA experiment in \cite{wan2021unsupervised}, with a batch size of 32 and learning rate of $5e-3$. We perform unsupervised model selection with the reconstruction loss of DIORA on the validation set. Our cascading and direct systems are trained for 10 epochs and 2000 batches respectively, as we found our direct systems to converge much faster. Further details are available in our training code\footnote{\href{https://github.com/roger-tseng/speech-parsing}{https://github.com/roger-tseng/speech-parsing}}.

\subsection{Evaluation}
\label{ssec:subhead}
Unsupervised constituency parsing on text is typically evaluated with $F_1$ score of constituents, where a match is only counted if a predicted constituent and a oracle constituent consist of the exact same words. However, erroneous word segmentation or ASR may introduce mismatch in the number of word-level leaves between model predictions and ground truth parse trees. Therefore, we match the constituents first by calculating an alignment between our word-level segments and oracle text, similar to SParseval \cite{roark2006sparseval}. 

We use forced alignment to determine the spans of audio that correspond to each word in the oracle sentence. 
We then compute the optimal one-to-one mapping that maximizes total span overlap between oracle segmentation and our proposed word segmentations\footnote{This mapping is determined via bipartite weight mapping, where the nodes are speech segments and the weights are given by the overlap duration across nodes.}. 
This allows us to first match nodes between predicted and ground truth parse trees, and calculate an $F_1$ score that jointly considers segmentation and parsing performance. We include whole sentence spans in our evaluation, in order to compare to AV-NSL \cite{anonymous2023textless}.
For all experiments, we evaluate fully unsupervised parsing with this proposed $F_1$ score, and report the average and standard deviation of corpus-level $F_1$ of the best model before convergence over five different random seeds. 
\subsection{Results of Cascading Systems}
\label{ssec:subhead}

We train two unsupervised ASR models to observe how varying accuracy of ASR may affect parsing performance.
The two models are trained with and without self-training following the original setup of wav2vec-U. 
They are denoted as ASR-ST and ASR respectively.
We use a 100-hour subset of speech from the training set, and 150k unpaired text sentences as our training data. 
Word-level transcriptions for the entire SpokenCOCO dataset are decoded from the phoneme output sequences of the ASR models. Word error rate of training set transcripts is 13.15\% and 28.25\% for AST-ST and ASR.

The training set transcripts are then used to train our parser. We follow \cite{zhao2020visually} and use the 10,000 most commonly occurring words in their respective training set transcripts as the vocabulary set for the parser.
Since we assume the availability of unpaired text data in the cascade scenario, we also consider training a parser from the unpaired text data alone. 


\begin{table}[htb]
\centering
\renewcommand{\arraystretch}{1.1}
\begin{tabular}{lcccc}
\toprule
\multirow{2}{*}{} & \multirow{2}{*}{Training split} & \multirow{2}{*}{Training} & \multirow{2}{*}{Testing} & \multirow{2}{*}{$F_1$} \\ 
\\
\midrule
\midrule
(A) & Entire train set        & oracle & oracle & 57.15 $\pm$ 2.09 \\
\midrule
(B) & Unpaired text           & oracle & ASR-ST & 44.08 $\pm$ 1.64 \\
(C) & Entire train set        & ASR-ST & ASR-ST & 40.53 $\pm$ 1.65 \\ 
\midrule
(D) & Unpaired text           & oracle & ASR & 34.97 $\pm$ 1.32 \\
(E) & Entire train set        & ASR & ASR & 31.01 $\pm$ 1.17 \\ 
\bottomrule
\end{tabular}
\caption{$F_1$ score of our casacading systems. 
The leftmost column lists whether training data comes from the unpaired text split or the training split.
We also list the $F_1$ score obtained by training and testing our parser on oracle transcripts in row (A) as a topline.
}
\label{table:1}
\end{table}


\vspace{1mm} \noindent \textbf{Effect of ASR errors on parsing.}
When comparing results across different blocks, we can see that parsing accuracy is heavily degraded when ASR errors are present. Additionally, one might expect that training a parser on ASR transcripts would allow it to better handle text with ASR errors during inference. However, by comparing rows (B)/(C) and rows (D)/(E), we see that training our parser on unpaired oracle text is consistently better than training on ASR transcripts. 
We note that this occurs in spite of the training set being nearly 3x larger than the unpaired text set. 
We hypothesize that this is partially caused by the decoding process of wav2vec-U. Uncommon words rarely get decoded by the WFST language model. As a result, the ASR transcripts contain fewer types of words compared to the original vocabulary, and parser trained on imperfect transcripts deal with more out-of-vocabulary words. The training set transcripts from the AST-ST model only use 8.2k words out of the original 16k words present in the ground truth captions.
\subsection{Results of Direct Systems}
\label{ssec:subhead}

For direct systems, only speech features and word boundaries are required. 
We extract frame-level speech features from the 14\textsuperscript{th} layer of XLSR-53 \cite{conneau21_interspeech}, a publicly available wav2vec 2.0 model pretrained on 53 languages.
For word boundaries, we naively split all utterances into 0.5-second segments, to encompass approximately one word in each segment\footnote{As a reference, average word length in SpokenCOCO is about 0.4 seconds, see Appendix of \cite{hsu-etal-2021-text}.}.

Since this method of segmentation is very inaccurate, the parsing results are similarly poor (Table~\ref{table:2} row (D)).
However, when provided with ground truth segmentation during testing (Table~\ref{table:2} row (C)), the parser trained on fixed length segments is able to achieve a performance similar to the parser trained on ground truth. 
This suggests that our direct system is limited by segmentation accuracy during inference. 

\vspace{-2mm}
\subsection{A Hybrid Approach: Segmenting speech with word boundaries determined by unsupervised ASR}
\label{ssec:subhead}
\begin{table}[htb]
\renewcommand{\arraystretch}{1.1}
\centering
\setlength\tabcolsep{4pt} 
\begin{tabular}{lcccc}
\toprule
\multirow{2}{*}{} & \multirow{2}{*}{Approach} & \multicolumn{2}{c}{Segmentation} & \multirow{2}{*}{$F_1$} \\ 
& & Training & Testing \\
\midrule
\midrule
(A) & AV-NSL & ground truth & ground truth     & 55.51 \\
(B) & Ours    & ground truth & ground truth     & 57.11 $\pm$ 0.00 \\
\midrule
(C) & Direct & every 0.5 sec. & ground truth   & 57.10 $\pm$ 0.01 \\
(D) & Direct & every 0.5 sec. & every 0.5 sec. & 3.88 $\pm$ 0.00 \\
\midrule
(E) & Hybrid & AST-ST       & AST-ST       & 40.44 $\pm$ 1.72 \\
(F) & Hybrid & ASR          & ASR          & 28.49 $\pm$ 0.57 \\
\bottomrule
\end{tabular}
\caption{$F_1$ score of direct and hybrid systems.
We include $F_1$ scores obtained by training and testing using oracle segmentation in rows (A) and (B) as toplines.
}
\label{table:2}
\end{table}

\noindent 
We compare our systems with AV-NSL under oracle segmentation settings in rows (A) and (B).
We find that our parser outperforms AV-NSL despite not using any visual grounding information.
This suggests that the DIORA framework may be better suited for unsupervised spoken constituency parsing.

We experimented with a speech-only unsupervised word segmentation method \cite{fuchs22_interspeech}, but found it to be suboptimal. Therefore, we consider a hybrid approach that uses forced alignment to obtain word boundaries from unsupervised ASR transcripts.
We find that when word boundaries are sufficiently accurate, using word boundaries alone can achieve similar accuracy to cascading systems, as shown in Table~\ref{table:1} row (C) and Table~\ref{table:2} row (E).
This implies that accurate word segmentation is necessary for unsupervised constituency parsing from speech, which aligns with the findings in AV-NSL.



\vspace{-3mm}
\subsection{On the Inductive Bias and Trivial Tree Structure}
\label{ssec:subhead}
Due to the head-initial property of English \cite{baker2008atoms}, constituency parse trees tend to be right-branching, especially if punctuation is removed.
On the other hand, for head-final languages such as Japanese and Korean, trees are left-branching instead.

In our direct and hybrid systems, we observe that our models tend to converge to producing right-branching trees on SpokenCOCO\footnote{We note that right-branching is a difficult baseline even for parsers trained on oracle text. As shown in Table~\ref{table:1} row (A) and \cite{wan2021unsupervised}, unsupervised text parsers only marginally improve on right-branching trees.}. 
It is worthwhile to note that our framework does not apply any inductive bias that encourages the model to favor right-branching trees; hence, it is non-trivial for such a phenomenon to emerge. 
We hypothesize that our systems learn a language's branching direction from continuous spoken input without supervision. 

\begin{figure}[htb]

\begin{minipage}[b]{1.05\linewidth}
  \centering
  \centerline{\includegraphics[width=8.0cm]{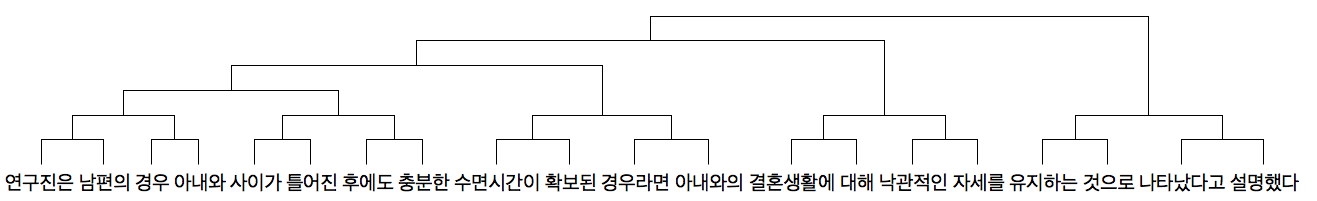}}
  \centerline{(a) Generated parse tree}\medskip
\end{minipage}
\hfill
\begin{minipage}[b]{1.05\linewidth}
  \centering
  \centerline{\includegraphics[width=8.0cm]{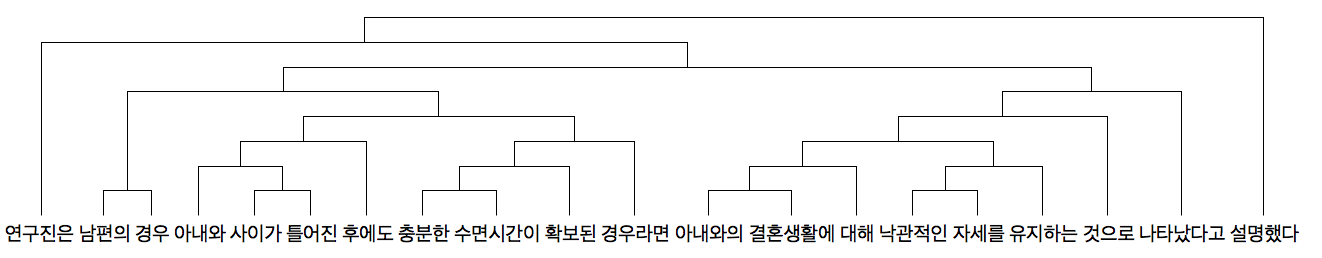}}
  \centerline{(b) Oracle parse tree}\medskip
\end{minipage}
\vspace{-6mm}
\caption{A sample pair of oracle and generated Korean parse trees. Only textual transcripts are shown for ease of visualization.}
\label{fig:res}
\end{figure}

We empirically verify this claim by conducting experiments on Korean, a primarily left-branching language. 
Over 5 runs with different random seeds, 3 runs converge towards producing some left-branching structures (Fig.~\ref{fig:res}),  supporting our hypothesis. 

\begin{table}[htb]
\centering
\begin{tabular}{cccc}
\toprule
\multirow{2}{*}{} & \multirow{2}{*}{English} & \multirow{2}{*}{Korean}  \\ 
\\
\midrule
\midrule
\multicolumn{1}{l}{\textit{Rule-based}} \\
Left branching              & 24.68 & 27.15 \\ 
Right branching             & 57.11 & 7.60 \\ 
\midrule 
\multicolumn{1}{l}{\textit{Speech only}} \\
0.5 sec. segmentation       & 57.10 $\pm$ 0.01 & 18.53 $\pm$ 8.99 \\
\bottomrule
\end{tabular}
\caption{Results of direct systems trained on English (right-branching) and Korean (left-branching), respectively, using the same setting as Table~\ref{table:2} row (C).}
\label{table:r}
\end{table}
\vspace{-3mm}



\vspace{-6mm}

\section{CONCLUSION}
\label{sec:conclusioni}
The work investigates cascading and direct approaches to perform constituency parsing on speech input, while only requiring raw speech and unpaired data. 
For cascading systems, we empirically show that parsers trained on ASR transcripts do not parse ASR transcripts better than parsers trained on unpaired text.
For direct and hybrid systems, our results suggest that using segmentation alone may be sufficient to produce unsupervised parse trees.

For future work, we expect to extend our system to end-to-end training to jointly optimize word segmentation and parsing. 
Additionally, we also plan to investigate whether unsupervised spoken constituency parsing can be improve other speech processing tasks under low-resource scenarios, such as text-to-speech, spoken question answering, or spoken content retrieval.

\vspace{2mm} 
\noindent
\textbf{Acknowledgement.} 
We thank the helpful discussions with Freda Shi, Shang-Wen Li, Ali Elkahky, and Abdelrahman Mohamed. We also thank the National Center for High-performance Computing (NCHC) of National Applied Research Laboratories (NARLabs) in Taiwan for providing computational and storage resources.

\vfill\pagebreak
\small{
\bibliographystyle{IEEEbib}
\bibliography{strings,refs}

\begin{thebibliography}{10}

\bibitem{10.5555/864689}
G.~Carroll et~al.,
\newblock ``Two experiments on learning probabilistic dependency grammars from
  corpora,''
\newblock Tech. {R}ep., USA, 1992.

\bibitem{klein-manning-2004-corpus}
D.~Klein et~al.,
\newblock ``Corpus-based induction of syntactic structure: Models of dependency
  and constituency,''
\newblock in {\em ACL}, 2004.

\bibitem{bod-2006-subtrees}
R.~Bod,
\newblock ``An all-subtrees approach to unsupervised parsing,''
\newblock in {\em COLING-ACL}, 2006.

\bibitem{shen2018neural}
Y.~Shen et~al.,
\newblock ``Neural language modeling by jointly learning syntax and lexicon,''
\newblock in {\em ICLR}, 2018.

\bibitem{strubell2018linguistically}
E.~Strubell et~al.,
\newblock ``Linguistically-informed self-attention for semantic role
  labeling,''
\newblock in {\em EMNLP}, 2018.

\bibitem{kuncoro2020syntactic}
A.~Kuncoro et~al.,
\newblock ``Syntactic structure distillation pretraining for bidirectional
  encoders,''
\newblock {\em TACL}, 2020.

\bibitem{Guo2019}
H.~Guo et~al.,
\newblock ``{Exploiting Syntactic Features in a Parsed Tree to Improve
  End-to-End TTS},''
\newblock in {\em Interspeech}, 2019.

\bibitem{tyagi20_interspeech}
S.~Tyagi et~al.,
\newblock ``{Dynamic Prosody Generation for Speech Synthesis Using
  Linguistics-Driven Acoustic Embedding Selection},''
\newblock in {\em Interspeech}, 2020.

\bibitem{song2021}
C.~Song et~al.,
\newblock ``Syntactic representation learning for neural network based tts with
  syntactic parse tree traversal,''
\newblock in {\em ICASSP}, 2021.

\bibitem{shen2018ordered}
Y.~Shen et~al.,
\newblock ``Ordered neurons: Integrating tree structures into recurrent neural
  networks,''
\newblock in {\em ICLR}, 2019.

\bibitem{kim2019unsupervised}
Y.~Kim et~al.,
\newblock ``Unsupervised recurrent neural network grammars,''
\newblock in {\em NAACL-HLT}, 2019.

\bibitem{drozdov-etal-2019-unsupervised-latent}
A.~Drozdov et~al.,
\newblock ``Unsupervised latent tree induction with deep inside-outside
  recursive auto-encoders,''
\newblock in {\em NAACL-HLT}, 2019.

\bibitem{zhu-etal-2020-return}
H.~Zhu et~al.,
\newblock ``The return of lexical dependencies: Neural lexicalized {PCFG}s,''
\newblock {\em TACL}, 2020.

\bibitem{yang-etal-2021-neural}
S.~Yang et~al.,
\newblock ``Neural bi-lexicalized {PCFG} induction,''
\newblock in {\em ACL-IJCNLP}, 2021.

\bibitem{shi-etal-2019-visually}
H.~Shi et~al.,
\newblock ``Visually grounded neural syntax acquisition,''
\newblock in {\em ACL}, 2019.

\bibitem{zhao2020visually}
Y.~Zhao et~al.,
\newblock ``Visually grounded compound pcfgs,''
\newblock in {\em EMNLP}, 2020.

\bibitem{wan2021unsupervised}
B.~Wan et~al.,
\newblock ``Unsupervised vision-language grammar induction with shared
  structure modeling,''
\newblock in {\em ICLR}, 2021.

\bibitem{anonymous2023textless}
C.~I. Lai et~al.,
\newblock ``Textless phrase structure induction from visually-grounded
  speech,'' 2023.

\bibitem{charniak2001edit}
E.~Charniak and M.~Johnson,
\newblock ``Edit detection and parsing for transcribed speech,''
\newblock in {\em NAACL}, 2001.

\bibitem{honnibal2014joint}
M.~Honnibal et~al.,
\newblock ``Joint incremental disfluency detection and dependency parsing,''
\newblock {\em TACL}, 2014.

\bibitem{kahn2005effective}
J.~G. Kahn et~al.,
\newblock ``Effective use of prosody in parsing conversational speech,''
\newblock in {\em HLT/EMNLP}, 2005.

\bibitem{dreyer07_interspeech}
M.~Dreyer et~al.,
\newblock ``{Exploiting prosody for PCFGs with latent annotations},''
\newblock in {\em Interspeech 2007}, 2007.

\bibitem{yoshikawa-etal-2016-joint}
M.~Yoshikawa et~al.,
\newblock ``Joint transition-based dependency parsing and disfluency detection
  for automatic speech recognition texts,''
\newblock in {\em EMNLP}, 2016.

\bibitem{pupier22_interspeech}
A.~Pupier et~al.,
\newblock ``{End-to-End Dependency Parsing of Spoken French},''
\newblock in {\em Interspeech}, 2022.

\bibitem{GROSJEAN197958}
F.~Grosjean et~al.,
\newblock ``The patterns of silence: Performance structures in sentence
  production,''
\newblock {\em Cognitive Psychology}, 1979.

\bibitem{price1991use}
P.~Price et~al.,
\newblock ``The use of prosody in syntactic disambiguation,''
\newblock in {\em Workshop on Speech and Natural Language}, 1991.

\bibitem{huang-harper-2010-appropriately}
Z.~Huang et~al.,
\newblock ``Appropriately handled prosodic breaks help {PCFG} parsing,''
\newblock in {\em NAACL-HLT}, 2010.

\bibitem{tran2018parsing}
T.~Tran et~al.,
\newblock ``Parsing speech: a neural approach to integrating lexical and
  acoustic-prosodic information,''
\newblock in {\em NAACL-HLT}, 2018.

\bibitem{ghaly2020using}
H.~Ghaly et~al.,
\newblock ``Using prosody to improve dependency parsing,''
\newblock in {\em International Conference on Speech Prosody}, 2020.

\bibitem{tran21_interspeech}
T.~Tran et~al.,
\newblock ``{Assessing the Use of Prosody in Constituency Parsing of Imperfect
  Transcripts},''
\newblock in {\em Interspeech}, 2021.

\bibitem{lakhotia-etal-2021-generative}
K.~Lakhotia et~al.,
\newblock ``On generative spoken language modeling from raw audio,''
\newblock {\em TACL}, 2021.

\bibitem{nguyen2020zero}
T.~A. Nguyen et~al.,
\newblock ``The zero resource speech benchmark 2021: Metrics and baselines for
  unsupervised spoken language modeling,''
\newblock in {\em NeurIPS SAS Workshop}, 2020.

\bibitem{dunbar2021zero}
E.~Dunbar et~al.,
\newblock ``The zero resource speech challenge 2021: Spoken language
  modelling,''
\newblock {\em arXiv}, 2021.

\bibitem{10.5555/1214993}
D.~Jurafsky et~al.,
\newblock {\em Speech and Language Processing (2nd Edition)},
\newblock Prentice-Hall, Inc., USA, 2009.

\bibitem{kasami1966efficient}
T.~Kasami,
\newblock ``An efficient recognition and syntax-analysis algorithm for
  context-free languages,''
\newblock {\em Coordinated Science Laboratory Report}, 1966.

\bibitem{younger1967recognition}
D.~H Younger,
\newblock ``Recognition and parsing of context-free languages in time n3,''
\newblock {\em Information and control}, 1967.

\bibitem{baevski2021unsupervised}
A.~Baevski et~al.,
\newblock ``Unsupervised speech recognition,''
\newblock {\em NeurIPS}, 2021.

\bibitem{bhati21_interspeech}
S.~Bhati et~al.,
\newblock ``{Segmental Contrastive Predictive Coding for Unsupervised Word
  Segmentation},''
\newblock in {\em Proc. Interspeech}, 2021.

\bibitem{kamper2022word}
H.~Kamper,
\newblock ``Word segmentation on discovered phone units with dynamic
  programming and self-supervised scoring,''
\newblock {\em arXiv preprint arXiv:2202.11929}, 2022.

\bibitem{hsu-etal-2021-text}
W.~N. Hsu et~al.,
\newblock ``Text-free image-to-speech synthesis using learned segmental
  units,''
\newblock in {\em ACL-ICJNLP}, 2021.

\bibitem{lin2014microsoft}
T.~Y. Lin et~al.,
\newblock ``Microsoft coco: Common objects in context,''
\newblock in {\em ECCV}, 2014.

\bibitem{kitaev-klein-2018-constituency}
N.~Kitaev et~al.,
\newblock ``Constituency parsing with a self-attentive encoder,''
\newblock in {\em ACL}, 2018.

\bibitem{mcauliffe17_interspeech}
M.~McAuliffe et~al.,
\newblock ``{Montreal Forced Aligner: Trainable Text-Speech Alignment Using
  Kaldi},''
\newblock in {\em Interspeech}, 2017.

\bibitem{roark2006sparseval}
B.~Roark et~al.,
\newblock ``Sparseval: Evaluation metrics for parsing speech,''
\newblock in {\em LREC}, 2006.

\bibitem{conneau21_interspeech}
A.~Conneau et~al.,
\newblock ``{Unsupervised Cross-Lingual Representation Learning for Speech
  Recognition},''
\newblock in {\em Interspeech}, 2021.

\bibitem{fuchs22_interspeech}
T.~Fuchs et~al.,
\newblock ``{Unsupervised Word Segmentation using K Nearest Neighbors},''
\newblock in {\em Interspeech}, 2022.

\bibitem{baker2008atoms}
M.~C. Baker,
\newblock {\em The atoms of language: The mind's hidden rules of grammar},
\newblock Basic books, 2008.

\end{thebibliography}
}

\end{document}